\documentclass[sn-mathphys,Numbered]{sn-jnl}% Math and Physical Sciences Reference Style
\usepackage{graphicx}%
\usepackage{multirow}%
\usepackage{amsmath,amssymb,amsfonts}%
\usepackage{amsthm}%
\usepackage{mathrsfs}%
\usepackage[title]{appendix}%
\usepackage{xcolor}%
\usepackage{textcomp}%
\usepackage{manyfoot}%
\usepackage{booktabs}%
\usepackage{algorithm}%
\usepackage{algorithmicx}%
\usepackage{algpseudocode}%
\usepackage{listings}%
\usepackage{lineno}%
\usepackage{blindtext}%
\usepackage{booktabs}
\usepackage{tablefootnote}
\usepackage{orcidlink}
\usepackage{ulem}
\usepackage{caption}
\usepackage{subcaption}
\newcommand{\OldNewText}[2]{{#2}\color{black}}

\newcommand{\BP}{Bridge-Prompt}

\begin{document}

\title[Article Title]{Zero-shot Prompt-based Video Encoder for Surgical Gesture Recognition}


\author[1]{\fnm{Mingxing} \sur{Rao}\orcidlink{0009-0006-6582-4598}}\email{mingxing.rao@vanderbilt.edu}

\author[1]{\fnm{Yinhong} \sur{Qin}\orcidlink{0009-0000-4526-4139}}\email{yinhong.qin@vanderbilt.edu}

\author[1]{\fnm{Soheil} \sur{Kolouri}\orcidlink{0000-0001-8495-5362}}\email{soheil.kolouri@vanderbilt.edu}
\author[1]{\fnm{Jie Ying} \sur{Wu}\orcidlink{0000-0002-7306-8140}}\email{jieying.wu@vanderbilt.edu}

\author*[1]{\fnm{Daniel} \sur{Moyer}\orcidlink{0000-0003-4428-5012}}\email{daniel.moyer@vanderbilt.edu}

\affil[1]{\orgdiv{Department of Computer Science}, \orgname{Vanderbilt University}, 
\orgaddress{\city{Nashville}, \country{USA}}}

% \affil[2]{\orgdiv{Department}, \orgname{Organization}, \orgaddress{\street{Street}, \city{City}, \postcode{10587}, \state{State}, \country{Country}}}

% \affil[3]{\orgdiv{Department}, \orgname{Organization}, \orgaddress{\street{Street}, \city{City}, \postcode{610101}, \state{State}, \country{Country}}}

% \linenumbers
\abstract{
\textbf{Purpose}: \OldNewText{Surgical video is an important data stream for gesture recognition. Thus, robust visual encoders for those data-streams is similarly important.}{
In order to produce a surgical gesture recognition system that can support a wide variety of procedures, either a very large annotated dataset must be acquired, or fitted models must generalize to new labels (so called ``zero-shot'' capability). In this paper we investigate the feasibility of latter option.
%Surgical gesture recognition plays a crucial role in the advancement of surgical automation.
%Given the significant variability in gesture vocabularies across different procedures, surgical videos serve as a critical data source for this recognition process.
%However, the lack of annotations in the vast majority of surgical videos presents a challenge.
%Consequently, developing a surgical gesture recognition system with zero-shot learning capabilities, which can generalize across diverse procedures without requiring extensive retraining, is essential.
}\\
\textbf{Methods}: Leveraging the Bridge-Prompt framework, we \OldNewText{fine-tune}{prompt-tune } a pre-trained vision-text model (CLIP) for gesture recognition in surgical videos. This can utilize extensive outside video data such as text, but also make use of label meta-data and weakly supervised contrastive losses.\\
\textbf{Results}: Our experiments show that prompt-based video encoder outperforms standard encoders in surgical gesture recognition tasks. Notably, it displays strong performance in zero-shot scenarios, where gestures/tasks that were not provided during the encoder training phase are included in the prediction phase. Additionally, we measure the benefit of inclusion text descriptions in the feature extractor training schema.\\
\textbf{Conclusion} Bridge-Prompt and similar pre-trained+\OldNewText{fine-tuned}{prompt-tuned } video encoder models present significant visual representation for surgical robotics, especially in gesture recognition tasks. Given the diverse range of surgical tasks (gestures), the ability of these models to zero-shot transfer without the need for any task (gesture) specific retraining makes them invaluable.
}




\maketitle

\section{Introduction}\label{sec1}

%The rapid evolution of computer-assisted intervention (CAI) in the surgical arena has presented opportunities to enhance surgical precision, safety, and outcomes. Central to these advancements is the role of surgical gesture recognition.

\OldNewText{Proposed intra-operative robotic gesture recognition systems are described as enabling real-time feedback, assisting in multi-arm guidance, and facilitating automated sub-tasks.}{
Proposed intra-operative robotic gesture recognition systems are described as enabling automation, rapid skill assessment and pedagogic feedback, and more general surgical support \cite{van2021gesture}.
}
However, current methods \cite{i3d,tran2015learning,spatiocnnResnet0} for gesture recognition attempt to estimate a fixed set of supervised gestures from contemporaneously collected video and/or kinematic datastreams \cite{symmetric_dilated_conv_recog, van2022gesture, sd-net, bounded_ms_tcn++}.
These systems are often trained in a fully supervised manner, which requires frame-wise gesture labels for the entire training set, and importantly require specification of the target labels during the whole training phase.
%Moreover are not built for generalization to unseen tasks and gestures.
%Surgical video, as one of the most important components in surgical dataset, drew lots of attention from researchers.
%Current robotic gesture recognition methods attempt to estimate specific gestures from frames of 
%While many studies combine these data with robotic kinematic information, video data offer rich information about surgical gestures (surgemes) and is crucial for clinical deployment of surgical recognition systems.
%The expressiveness of the visual features directly influences the discriminating power of any gesture-recognition system. A common practice is to transplant image encoder architectures (visual feature extractors) from natural image tasks. Examples include Inflated-3D (I3D) \cite{i3d}, 3DCNN\cite{tran2015learning}, 3DResNet\cite{spatiocnnResnet0}. Training these encoders is often done in a fully supervised manner, requiring frame-wise gesture labels, a costly and error-prone resource \cite{van2019weakly}. 

%\BlueComment{Critical Sentance Goes Here?}
%\OldNewText{As procedures vary greatly in gesture vocabulary, }{T}rained recognition systems will need the capacity to generalize without extensive retraining \OldNewText{}{or large supervised datasets in order to cover };
\OldNewText{As procedures vary greatly in gesture vocabulary, a trained recognition systems will need the capacity to generalize without extensive retraining}{We believe that in order to produce a system that can provide support to wide variety of procedures, gesture recognition as a sub-field will either need a) a large number of annotated datasets for each procedure as previous fully supervised methods demand, or b) a system that can generalize well to new label sets with limited supervision.
The latter option is likely more efficient given the expense of annotation, and fits within the ``zero-shot learning'' paradigm, where a general representation  is reused for labelling tasks unseen during training \cite{wang2019survey}.
}

%\OldNewText{}{In order to demonstrate the feasibility of a zero-shot scheme,}
% \BlueComment{Ethan, this section will need work, but I have rewritten it in a ``Bridge-Prompt First'' manner}
\OldNewText{In the present work we}{We thus } investigate the feasibility of such a zero-shot gesture recognition method using weak supervision and text augmentation\OldNewText{ for pre-training}{}.
We focus on improving visual feature extraction from video data streams, as these data offer rich information about surgical gestures (surgemes) and kinematic data require access to the research API, limiting current use cases.
%and we believe that video derived features transfer across task and domain relatively well in comparison with kinematic data, which may be machine calibration dependent and cannot be accessed outside of a research API.
We use the Bridge-Prompt \cite{bridge-prompt} framework
%, which transfers a pre-trained vision-text model (CLIP \cite{clip}) to an gesture classification specific domain.
which specifically relaxes the fully supervised constraints:
\BP{} is able to use large weakly supervised datasets, which are relatively inexpensive and numerous in comparison to densely annotated data, and our experiments suggest that \BP{} generalizes well to unseen labels \OldNewText{outside of pre-training phase}{}(i.e., novel gestures).

In order to validate these claims and evaluate \BP's overall efficacy and zero-shot capability, we demonstrate usage on the JIGSAWS \cite{jigsaws} and RARP-45 \cite{van2022gesture} dataset, simulating both standard and zero-shot use cases on cross-validated training schemes. We \OldNewText{fine-tune an encoder contrastively to the training fold data}{compare image encoders with differing configurations}, varying experimentally the amount of label information (both annotation presence/absence and label text). We then compare the performance of experimental cases and standard baselines from the literature using a standard prediction recognition model (MS-TCN++, \cite{mstcn++}). Our experiments show the benefit of using \BP{} image encoders for gesture recognition tasks.

%This provides relatively strong zero-shot performance, as we show empirically.

In summary, in the present work, we do the following:
\vspace{-2mm}
\begin{itemize}
%\item We demonstrate the use of Bridge-Prompt \cite{bridge-prompt} for pre-training surgical gesture recognition video encoders, showing that this method has superior performance compared to baseline encoders on a standard dataset \cite{jigsaws}.
%\item We further demonstrate zero-shot capability from Bridge-Prompt, wherein strong experimental performance is achieved on tasks that were not provided to the encoder at training time.
%\item We show the potential value of the gesture labels' text descriptions within training; these meta-data are ignored in other image feature extractors. We measure their marginal benefit, in comparison to null-case text descriptors.
%\item All of our experimental code is open source and is available at \url{https://github.com/mx-ethan-rao/Robotic_Gesture_Tools}.
%\item We show that using Bridge-Prompt \cite{bridge-prompt} for pre-training encoders improves surgical gesture recognition
\item We demonstrate the within-task and zero-shot capability of Bridge-Prompt 
\item We show that gesture labels' text descriptions do not improve training 
\item We provide open-source code\footnote{\url{https://github.com/mx-ethan-rao/Robotic_Gesture_Tools}} for \OldNewText{pre-training}{prompt-tuning } encoders with Bridge-Prompt
\end{itemize}

\paragraph{Terminology}

\OldNewText{}{Throughout this manuscript we use the phrase ``\emph{pre-trained}'' exclusively to denote previously trained instances of encoder networks that we either directly use without modifying or further train using a different weakly-supervised loss function. We refer to this further training phase as ``\emph{prompt tuning}''.

No matter if an encoder is only pre-trained or incorporates prompt-training, its weights are then frozen and used unchanged for the \emph{supervised training} phase, wherein a supervised network will be trained to predict gesture labels from the frozen embeddings.}

\section{Related Work}\label{sec2} 

Surgical robotic gesture recognition has an extensive history of study \cite{dipietro2016recognizing, tao2013surgical, reiley2008automatic}.
%\CiteLater{all citations from that review paper}.
While multiple disparate modalities are each reasonable to incorporate into a recognition system, video is by far the most popular \cite{sd-net,symmetric_dilated_conv_recog,3dconv_recog,stcnn_gd}, followed by integration of video with robotic kinematic data \cite{gesture_classification_kv_2013, graph_kv_embed_wu, van2022gesture}, and discrete surgical event streams \cite{davincinet}.
Notably, almost all proposed methods rely on video data, or on video derived features such as optical flow \cite{cross_modal_wu}.
%Derived features of imaging \cite{cross_modal_wu} and \cite{optical_flow_recog} also use optical flow extracted from surgical videos and has proved its ability to preserve core information under different backgrounds.

% \BlueComment{Talk about prediction methods here: } 
Gesture recognition is usually considered in a temporal context, and thus many of the prediction models are taken from time-series prediction and forecasting domains: HMM, LSTMs, temporal convolution, and attention methods \cite{tao2012sparse, dipietro2019segmenting, tcn, van2022gesture}.
%Relational graph learning \cite{graph_kv_embed_wu} and multi-modal attention mechanisms \cite{van2022gesture} represent modern techniques to capture latent joint knowledge between surgical video and kinematics data. 
The Temporal Convolutional Network (TCN) has emerged as a common selection for the majority of deep learning based temporal gesture studies \cite{tcn, mstcn, mstcn++}. \OldNewText{}{MS-TCN and MS-TCN++ \cite{mstcn, mstcn++} are the most relevant instances of this family of temporal modeling methods, as they were specifically designed for temporal action segmentation, and have been used in several surgical workflow analysis tasks \cite{yuan2022anticipation, czempiel2020tecno}. }
%In the present work, we use the TCN as the default prediction method.
Our objective is to test the effectiveness of differing image encoders \OldNewText{}{and not the prediction head architectures}, thus we use a generic MS-TCN++ implementation throughout the experiments.%; while more specialized architectures may produce marginally better results overall (e.g. attention-based TCN as in van Amsterdam et al. \cite{van2022gesture}), for understanding the improvements of the image encoding portion of the method, we believe it is best to use a generic method.

Feature extraction has become an essential part of deep learning systems \cite{bengio2012unsupervised}, particularly in computer vision \cite{clip}. Initially performed by fully convolutional architectures trained in a supervised end-to-end schema, recent image encoders are typified by extended pre-training \cite{simclr} using proxy tasks and/or self-supervised, contrastive methods. In surgical video, standard methods include Inflated-3D\cite{i3d}, 3DCNN\cite{tran2015learning}, and 3DResNet\cite{spatiocnnResnet0}; these methods fall into the category of fully convolutional supervised end-to-end trained encoders. \BP{}, which uses the pre-trained CLIP \cite{clip} vision-text joint embedding model, has been proposed as a generic video encoder. The objective of this paper is to measure both Bridge-Prompt's performance on the standard surgical gesture task and its zero-shot generalization ability.

\paragraph{Zero-Shot Learning}\label{subsubsec3}

% \BlueComment{TODO: rewrite this about background material in zero-shot learning}

In traditional learning, models learn from labeled samples for each class and make predictions on previously unseen samples of the same classes. In zero-shot learning, models are trained on a subset of classes and tested on a different set of classes without any overlap \cite{wang2019survey}. This is especially crucial in surgical robot videos where collecting annotated gestures for every possible class is impractical or expensive, leading to a vast amount of surgical video being unlabeled.

\section{Method}\label{sec3}

%Throughout this manuscript we start with the weights from multiple ``pre-trained'' networks.

This section describes the two parts of our model: 1) the video encoder (Bridge-Prompt) and 2) the downstream gesture recognition model.
The former we construct during \OldNewText{pre-training}{prompt-tuning } and is the focus of our empirical study, while the latter is used in evaluation of each of the encoders, including both the proposed encoder (Bridge-Prompt) and baseline encoders. Though neither is first proposed by this manuscript, we include descriptions of both, as understanding their nuances (or simplicity, in the case of the downstream model) is necessary for contextualizing experiments.

%We train a video encoder and use the trained video encoder for the frame-wise visual embedding. Then, we use a trivial downstream gesture recognition model to test the capability of embedded visual features.

\paragraph{Bridge-Prompt \OldNewText{Pre-training}{Prompt-tuning } Architecture}\label{subsec_video_encoder}

Bridge-Prompt is a training protocol for constructing high-quality image encoders for sequential labelling of video frames. It starts with an image-text joint encoder, for which the standard model is the CLIP model \cite{clip}, which is in turn based on ViT-B/16 vision transformer trained on three large natural image datasets \cite{lin2014microsoft, krishna2017visual, thomee2016yfcc100m}, alongside an analogous text transformer to GPT-2 \cite{radford2019language}. In the CLIP training protocol, these two models were modified to have matching encodings, so that from either the image or text one could predict the other. \BP{} starts at this pre-trained CLIP state and prescribes additional video-sequence based training. This fine-tuning to surgical video (and to the particular surgical gesture labels) is the first phase of our empirical work.

\begin{figure}[t]%
\centering
\includegraphics[width=\textwidth, trim={0.5cm 0.1cm 0.1cm 0.1cm}, clip]{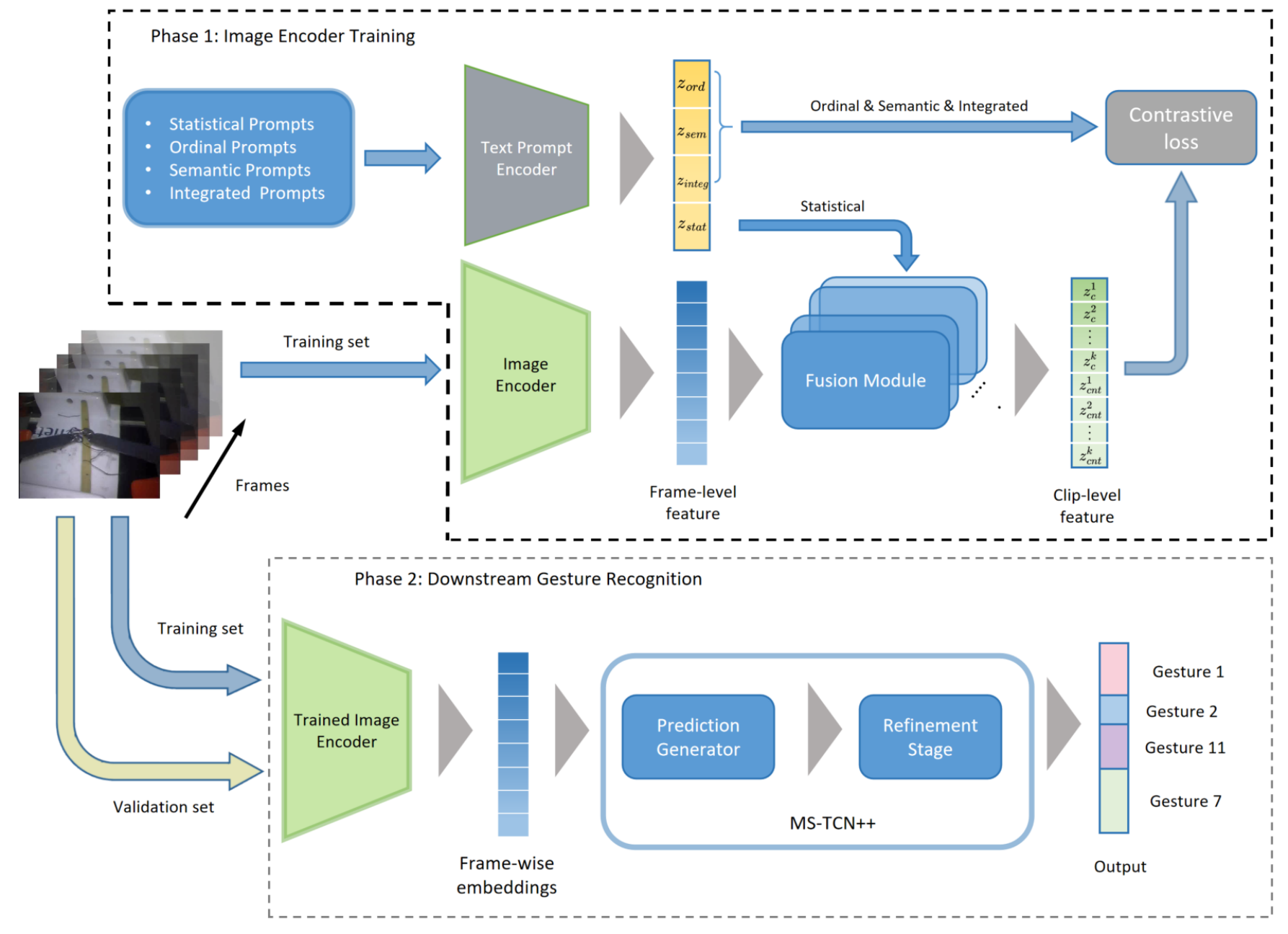}
\caption{The two phases of our training schema: at \textbf{top} the \BP{} pre-training, at \textbf{bottom} a ``simple probe'' predictor measuring performance on the supervised gesture recognition task.}\label{fig1}
\end{figure}

We follow the  \BP{} protocol and first split videos into sub-videos with a fixed number of frames, but possibly at different sampling rates. All resulting sub-videos have the same number of frames, alongside a label for each frame (which may be missing/undefined). Every defined label may also have a text description, e.g., ``orienting needle'' or ``pulling suture with left hand'', though this may also be left undefined, or, as we experiment with, replaced with a categorical placeholder.

For each sub-video four text prompts are constructed from the sub-video's labels:
\begin{enumerate}
\item {} \textbf{Statistical text prompt}: ``\textit{this video contains K actions in total}''
\item \textbf{Ordinal text prompt}: ``\textit{this is the i\textsuperscript{th} action in the video}'' (defined for each distinct label interval in the video)
\item \textbf{Semantic text prompt}: ``\textit{\{$Ord_i$\}, the person is performing \{Gesture $i$ text description\}}'' where \textit{$Ord_i$} refers to ``\textit{Firstly}''\footnote{The original \BP{} paper uses ``Firstly'' instead of ``First'', and we do not modify it here.}, ``\textit{Secondly}'', etc..
\item \textbf{Integrated text prompt}: the concatenation of all of the semantic text prompts.
\end{enumerate}
These prompts are then sent to the text encoder, to form $z_{stat}$, $z_{ord}$, $z_{sem}^k$, and $z_{int}$ respectively, where $k$ represents k-th gesture in a video clip.

Each image frame $x_t$ is passed through the initial image encoder $f$; this is the encoder that will be reused for the downstream surgical gesture recognition task. However, for fine-tuning only, the encodings $f(x_t)$ are then processed as a sequence by a ``fusion module'', which also receives as input the ordinal text prompts ($z_{ord}$) and summary statistic frame-level indicators (count tokens, split indicators, and length/position indicators). The outputs of the fusion module (the ``fusion encodings'') include $z_c^k$ for each gesture, mean-pooled $\bar{z}_c$, and a separate embedding $z_{count}$. $z_{count}^k$ is output at sub-clip for each gesture, but only the mean-pooled aggregate is used. These encodings are the focus of the loss components which drive the contrastive fine-tuning.

\paragraph{Contrastive Pre-training Losses}

%\subsubsection{Joint Video-Text Contrastive Learning}\label{subsubsection_video_text_contrastive}

%The ultimate goal of the Bridge-Prompt framework is to bridge the gap between video and text. The joint vision-text representation learning process aims to maximize the similarity between the encoded vision features and the textual descriptors. By leveraging the cosine similarity:
For two vectors $z_x,z_y$ on the same space the cosine similarity is
\begin{align}
\text{s}(\mathbf{z_x},\mathbf{z_y})=\frac{\mathbf{z_x} \cdot \mathbf{z_y}}{\|\mathbf{z_x}\|\|\mathbf{z_y}\|}
\end{align}
We can then define a batch similarity matrix from sets $Z_x = \{z_{x,b}\}$ and $Z_y = \{z_{y,b}\}$ 
\begin{align}
\text{S}(Z_x, Z_y) = 
\begin{bmatrix}
    \text{s}(z_{x1}, z_{y1}) & \dots & \text{s}(z_{x1}, z_{yB}) \\
    \vdots & \ddots & \vdots \\
    \text{s}(z_{xB}, z_{y1}) & \dots & \text{s}(z_{xB}, z_{yB})
\end{bmatrix}
\end{align}
where $b$ denotes the sub-video (batch) index and $B$ is the batch size. We construct three different similarity matrices: $S_{\text{sem}}^k =\text{S}(Z_{c}^k ,Z_{\text{sem}}^k)$ the similarity between the frame-wise encodings and the semantic text prompt embeddings, $S_{\text{int}} = \text{S}(\bar{Z}_c,Z_{\text{int}})$ the similarity between the mean-pooled embeddings and the integrated text prompt emeddings, and $S_{\text{stat}} = \text{S}(\bar{Z}_{count},Z_{\text{stat}})$ the similarity between the mean-pooled count embeddings and the statistical text prompt embeddings. After computing each similarity matrix soft-max is applied first row-wise/column-wise to form text-wise/clip-wise $\bar{S}_{\text{sem}},\bar{S}_{\text{int}}$, and $\bar{S}_{\text{stat}}$ respectively.

Within each batch, matching image-text pairs are taken as positive constrastive pairs, while mismatched pairs (i.e., videos paired with label text from a different video) are taken as negative contrastive pairs; this is to say that in the context of our contrastive learning problem we optimize the matrices towards the identity matrix. Towards this end we define three losses for each of the three matrices:
\begin{align}
\mathcal{L}_{\text{sem}}^k = \frac{1}{2}[\text{D}[ S_{sem}^k \| I] + \text{D}[ I \| S_{sem}^k]] ~~~~ ~~~~ \mathcal{L}_{\text{stat}} & = \frac{1}{2}[\text{D}[ S_{\text{stat}} \| I] + \text{D}[ I \| S_{\text{stat}}]] \label{eq:loss}\\
\mathcal{L}_{\text{int}} = \frac{1}{2}[\text{D}[ S_{\text{int}} \| I] + \text{D}[ I \| S_{\text{int}}]]\nonumber
\end{align}
where the (generalized KL) divergence $\text{D}$ is defined for square matrices of matching dimension
\begin{align}
\text{D}[A \| B] = \sum_{i=1}^N \sum_{j=1}^N A_{ij} \log \frac{A_{ij}}{B_{ij}}.
\end{align}

\paragraph{Gesture Recognition Model}\label{subsection_gesture_recog}
To measure the effectiveness of prompt-based video encoder, \OldNewText{}{during the evaluation } we freeze the weights in the video encoder and do frame-wise visual embedding. \OldNewText{}{The frame encoder is fixed, and } we train a predictive model for gesture recognition based on \OldNewText{the}{that fixed } visual embedding. \OldNewText{}{Our chosen downstream predictive model is the } MS-TCN++\cite{mstcn++}\OldNewText{; in order to ensure fair experiments, we use this same architecture across all experiments and all embedding methods, retraining it for each tested embedding. } Temporal Convolution Networks have become a common choice in action segmentation, and the MS-TCN++ is a refinement of the original MS-TCN. MS-TCN++ uses a simple two-stage training and a slightly modified convolutional configuration with dilations. We avoid deeper or more nuanced architectures (such as those incorporating attention, or longer context windows) to ensure that performance is due to the quality of the features and not the complexity of the classifier itself.

\section{Experiments}\label{section_experiments}
%Our experimental results show that this video encoder performs well not only in regular recognition tasks but also in zero-shot transferring tasks.

\paragraph{Datasets, Implementation, Training, \& Evaluation}\label{eval}

%JHU-ISI Gesture and Skill Assessment Working Set (JIGSAWS) \cite{jigsaws} represents a benchmark for gesture recognition. It is composed of both endoscopic video and robotic kinematic data of the da Vinci Surgical System (dVSS, Intuitive Surgical, Sunnyvale, CA, USA). JIGSAWS is collected from 8 surgeons with varying skill levels by performing surgical tasks, including suturing, knot-tying, and needle-passing, on phantom environments. Each video segment contains frame-level labels of its gesture type, which belongs to a predefined gesture dictionary of 15 different gestures. In our experiments, we only choose the video part of JIGSAWS and use the Leave-One-User-Out (LOUO) \cite{jigsaws} policy as our cross-validation method.

%To better illustrate our model’s generalizable performance, we also completed experiments on RARP-45 (Robot-Assisted Radical Prostatectomies) \cite{van2022gesture} dataset. RARP-45 is also collected from 8 surgeons with different surgical seniority using dVSS. Compared to JIGSAWS, it brings challenges since all video segments are captured during real surgery processes and the average length of each segment, around 292 seconds, is ~3 times as the average video length in JIGSAWS, around 92 seconds. RARP-45 has a gesture dictionary with a capacity of 8. See appendix for full comparison.

\textbf{Datasets:} We demonstrate \BP{} and baseline methods on two standard datasets: the JHU-ISI Gesture and Skill Assessment Working Set (\textbf{JIGSAWS}) \cite{jigsaws}, which is composed of endoscopic video of suturing and knot-tying in a phantom environment, and 
Robot-Assisted Radical Prostatectomies (\textbf{RARP-45}) \cite{van2022gesture}, which is composed of endoscopic video recordings of prostatectomies. Both datasets are collected from 8 surgeons with varying skill levels using the da Vinci Surgical System (dVSS, Intuitive Surgical, Sunnyvale, CA, USA), and are annotated for multiple gestures at the image-frame level: 15 gestures at 30\,Hz for JIGSAWS, and 8 gestures at 60\,Hz for RARP-45. JIGSAWS additionally has robotic kinematic recordings but we do not use these data in \BP{} trained methods. We also omit the JIGSAWS needle passing task due to data quality.
% We use the JHU-ISI Gesture and Skill Assessment Working Set (JIGSAWS) \cite{jigsaws} in our experiments. It is composed of both endoscopic video as well as robotic kinematic data of the da Vinci Surgical System (Intuitive Surgical, Sunnyvale, CA, USA), performing suturing, knot-tying, and needle-passing on phantom environments, and corresponding dense label annotations for those timeseries of each surgical gesture (with 15 different gestures including start and stop placeholders). We only use the video portion of the dataset for the knot-tying and suturing tasks, which leaves 103 videos with an average length of 90 seconds each, as well as their corresponding label sets and label descriptions. There are eight separate users of varying skill level present in the dataset, with the user in each video noted in the dataset metadata.

% For all of our experiments we use Leave-One-User-Out (LOUO) \cite{jigsaws} cross-validation. Importantly, for each of the pre-training stages we follow this experimental split as well, meaning that we train a video encoder on each training split before training the prediction network again on this training split, and only afterwards measuring the predictive accuracy on the test fold.

% RARP-45 dataset contains surgical video segments from Robot-Assisted Radical Prostatectomies (RARP) performed by 8 surgeons with different surgical seniority using the da Vinci Si Surgical System. These videos are captured at 60Hz and the frame size is 1920*1080. Each video includes gesture annotations from a dictionary of 7 fine-grained bimanual actions.

\textbf{Implementation: }We implement multiple \BP{} variants as well as two baseline image encoders (3DResNet \cite{spatiocnnResnet0} and I3D \cite{i3d}) in PyTorch. Each \BP{} implementation is composed of a backbone image encoder (either the default ViT-B/16 \cite{dosovitskiy2020image} or ResNet-50 \cite{he2016deep}) and \OldNewText{}{the analogous GPT-2 text encoder } \cite{radford2019language} which is discarded after training. These backbones are initialized using standard pre-trained weight-sets\footnote{Both \BP{} and 3DResNet methods prescribe the use of weights from another task and training phase before the \OldNewText{pre-training}{prompt-tuning } phase.\OldNewText{; in this paper ``pre-trained'' weights are from this first phase, while ``pre-training'' are the actions performed on gesture recognition data.}{}}.
%Two \BP{} variants exchange encoding architectures (RN50 \CiteNeeded and Vit-B/16 \CiteNeeded), but otherwise all of variations are due to changes in the \BlueComment{pre-}training dataset.
The \BP{} encoders are then \OldNewText{fine-tuned}{prompt-tuned } using the Adam optimizer minimizing the sum of the losses in Eq. \ref{eq:loss} in the \OldNewText{\textbf{pre-training}}{prompt-tuning } phase.

After \OldNewText{pre-training}{prompt-tuning } we freeze the weights of each image encoder, then train a standard prediction network for the supervised task (MS-TCN++ \cite{mstcn++}). This second directly supervised phase we call the \textbf{supervised training phase}. The \OldNewText{}{supervised } training phase has its own hold-out test set and it is on this hold-out that we report performance metrics.
For JIGSAWS we use Leave-One-User-Out (LOUO) \cite{jigsaws} cross-validation, and for RARP-45 we choose 10 videos (out of 36 total) as the test set.

%\BlueComment{Possible implementation details?}Vit-B/16 \cite{dosovitskiy2020image} is the default pre-trained image encoder in \BP{}. We enhanced \BP{} by integrating the RN50 \cite{he2016deep} image encoder, pre-trained in CLIP, to demonstrate its superiority. \BlueComment{Add MS-TCN++ details} Then we use MS-TCN++ \cite{mstcn++} to evaluate the encoded visual features.

%In the most basic experimental scheme we evaluate on the same within task (e.g., within the Knot-tying holdout set). We further train and evaluate on 

\textbf{Training samples:}
Even though we are constructing frame-by-frame image encoder, their training requires multi-frame segments (video clips) and their corresponding sequence of gesture labels.
Each video clip is sampled from the original video of JIGSAWS or RARP-45 in 16 frame windows at three separate temporal sampling rates (sampled frames every 4/8/16 frames for JIGSAWS and 6/15/30 frames for RARP-45). %with certain frequencies and overlaps.
We resize each frame to 224x224 pixels. The input format for the video clips for all methods (\BP{} \cite{bridge-prompt}, I3D \cite{i3d}, 3DResNet \cite{spatiocnnResnet0}) are the same, and input sets are changed only by the stated experimental condition. %The labels for them are given correspondingly.
For the label sequences from JIGSAWS, we provide two additional placeholder labels/prompts for the unlabelled frames at the beginning and end of each video: ``Waiting and preparing for the surgery'' for beginning frames, and ``Finishing the surgery'' for ending frames.

%In JIGSAWS, the beginning frames and ending frames are usually unlabeled. They are abandoned in I3D and 3DResNet. In our prompt-based video encoder, We provide prompts of ``Waiting and preparing for the surgery'' for beginning frames, and ``Finishing for the surgery'' for ending frames.

\textbf{Training time:} All experiments were run on NVIDIA A40 GPUs or NVIDIA A5000 GPUs. The \OldNewText{pre-training}{prompt-tuning } phase for all \BP{} variants was conducted on two GPUs, otherwise only a single GPU was used.
%We set up validation sets not only in downstream gesture recognition(phase 2) but also in video encoder training (phase 1).
In JIGSAWS, pre-training for 50 epochs for each tested variant of \BP{} takes approximately 8 hours using two A40 GPUs.
%We have 8 validation sets due to 8 users.
%For the frame-wise visual feature extraction, it takes one hour to process every frame in JIGSAWS original video.
%It then takes an additional hour to process the entire JIGSAWS dataset into encodings.
%For the downstream gesture recognition, it takes 5 minutes for one validation set.
It takes 5 minutes to train the MS-TCN++ during the supervised training phase, using pre-extracted image encodings.

\textbf{Performance metrics: }
We assess the outcomes using five standard evaluation metrics, Accuracy \OldNewText{}{(Acc.)}, Edit Distance, and F1@\{10,25,50\} \cite{tcn}. For JIGSAWS we condition this on pre-training task (Knot Tying or Suturing), and for zero-shot cases we also condition on each unseen label. %zero-shot generalization results we also report these 

%\subsection{Experiments}\label{subsec3}
%\paragraph{Visual Feature Quality}\label{subsubsec1}

\paragraph{Experimental Conditions}

We first measure the quality of the features learned by the various encoders operating under normal (non-zero-shot) conditions. Results are reported in Table \ref{tab1_1:encoders} and \ref{tab1_2:RARP-45}.
These results show that \BP{} improves gesture recognition performance as measured by all but one performance metric. Further, due to the similarity between \BP{} performance with either ResNet50 or ViT backbones, this performance gain does not appear to be due to the transformer architecture of ViT.
However, ResNet50 backbones appear to be slightly less stable in RARP-45 training; across multiple runs we experienced exploding gradients, and the method failed to converge. Tuning the learning-rate might resolve this issue, but we did not have the resources to tune that parameter.
We include JIGSAWS results reported in van Amsterdam et al. 2022 for contextualization\footnote{\label{foot:van}Van Amsterdam et al. 2022 \cite{van2022gesture} reports results for models that include input data from both video and robotic kinematic streams, and thus are not entirely comparable for selecting video encoders. Moreover, the particular architecture, training details, and data split cannot be reproduced without access to their codebase. We exclude their RARP-45 reported values as they evaluate on all 45 videos, but only 36 videos in RARP-45 are publicly available.},
as we believe this to be a state-of-the-art contemporary system utilizing all available data streams (both visual and kinematic), and (presumably) many model architecture optimizations and procedure refinements.
%though the results reported therein are for networks that also incorporate both the kinematic data, and more complex attention based prediction model.
%We devise three distinct experimental methodologies to show the performance of the prompt-based video encoder. Our evaluations focus on the quality of visual features, overall performance comparison for limited gesture training, zero-shot recognition capabilities for previously unseen gestures, and cross-task learning capabilities. We adopt Leave-One-User-Out (LOUO) \cite{jigsaws} cross-validation. \textbf{It is worth noticing that the validation set for the downstream recognition model is also excluded for the video encoder training. In other words, we train eight different video encoders for each experiment due to 8 users (surgeons) in JIGSAWS.}

%Inflated3D \cite{i3d}(I3D) model has been widely used in gesture recognition\cite{van2022gesture} and action segmentation\cite{mstcn}, \cite{mstcn++} as a video feature extractor. We conduct a comparative analysis of its visual feature attributes relative to those of the prompt-based video encoder.  

%\setcustomtablecounter{1}{1}
\begin{table}[t]
\captionsetup{font=footnotesize}
\caption{Here we present gesture recognition results on the JIGSAWS datatset for 3DResNet, I3D, and Bridge Prompt (BP) with both ResNet50 and ViT variants, as well as the Bridge Prompt architecture with no JIGSAWS specific \OldNewText{pre-training}{prompt-tuning} (CLIP).
%All methods were pre-trained on the same experimental splits, and all use the same simple TCN downstream prediction head.
We provide the MA-TCN[C] numbers reported by van Amsterdam et al. \cite{van2022gesture} for overall context, but MA-TCN[C] uses both the video and kinematic data streams and a more complex prediction model.}\label{tab1_1:encoders}

\begin{tabular*}{\textwidth}{@{\extracolsep\fill}lcccccc}
\toprule%
& \multicolumn{3}{@{}c@{}}{Knot Tying} & \multicolumn{3}{@{}c@{}}{Suturing} \\
\cmidrule{2-4}\cmidrule{5-7}%
  & Acc. & Edit & F1@\{10, 25, 50\} & Acc. & Edit & F1@\{10, 25, 50\} \\
\midrule
3DResNet  & 66.3 & 65.3 & 69.7, 65.56, 51.14  & 68.19 & 67.78 & 73.37, 69.06, 57.87\\ %
\midrule
I3D  & 68.39 & 74.36 & 78.20, 70.27, 54.21 & 68.63 & 75.93 & 78.44, 74.28, 59.69\\ %
\midrule
CLIP-ViT & 70.42 & 75.52 & 77.41, 72.52, 59.77 & 75.15 & 78.81 & 83.35, 79.94, 68.02\\ %
\midrule
CLIP-ResNet50  & 69.05 & 76.54 & 79.06, 73.21, 59.55 & 71.04 & 77.91 & 81.35, 77.28, 64.57 \\ %
\midrule
BP-ResNet50  & 77.82 & \textbf{78.97} & \textbf{84.79}, \textbf{81.03}, 66.39 & 81.42 & \textbf{84.71} & \textbf{89.12}, \textbf{87.71}, 77.60 \\ %
\midrule
BP-ViT  & \textbf{81.00} & 74.19 & 81.19, 78.58, \textbf{68.31} & \textbf{81.72} & 83.89 & 87.34, 85.57, \textbf{77.61}\\ %
\hline
\midrule
MA-TCN[C] & \multicolumn{3}{@{}c@{}}{ \tiny \color{red} Causal variant from \cite{van2022gesture}, see footnote \ref{foot:van}. }
& 83.4 & 81.6 & 87.7 {\tiny \color{red}~~only F1@10 given}
\\
%MA-TCN[A] & \multicolumn{3}{@{}l@{}}{ \tiny \color{red} Acausal variant from \cite{van2022gesture}, see footnote \ref{foot:van}. }
%& 86.8 & 91.4 & 94.3 {\tiny \color{red} only F1@10 given}
%\\
\botrule
\end{tabular*}
\end{table}

%\setcustomtablecounter{1}{2}
\begin{table}[h]
\captionsetup{font=footnotesize}
\caption{Here we present gesture recognition results on the RARP-45 datatset for 3DResNet, I3D, and Bridge Prompt (BP) with the ViT variant, as well as a ``no RARP-45 specific \OldNewText{pre-training}{prompt-tuning}'' case (CLIP). BP-ResNet50 failed to converge, thus we decide not to post it.
%All methods were pre-trained on the same experimental splits, and all use the same simple TCN downstream prediction head.
%using gesture indices (e.g., ``Gesture 1'', ``Gesture 2'') instead of gesture descriptions during \BP{} training.
}\label{tab1_2:RARP-45}
\begin{tabular*}{\textwidth}{@{\extracolsep\fill}lccccc}
\toprule%
  & Acc. & Edit & F1@10 & F1@25 & F1@50 \\
\midrule
3DResNet  & 66.97 & \textbf{76.95} & 71.76 & 66.34 & 52.90 \\ 
I3D  & 65.95 & 74.41 & 71.52 & 65.35 & 51.23 \\ 
CLIP-ViT & 70.00 & 73.88 & 74.94 & 73.30 & 59.95 \\ 
CLIP-ResNet50  & 68.93 & 71.85 & 70.83 & 66.90 & 55.09 \\  %
% BP-ResNet50  & - & - & - & - & - \\  %
BP-ViT  & \textbf{77.36} & 72.16 & \textbf{76.29} & \textbf{74.50} & \textbf{65.32} \\ %
\botrule
\end{tabular*}
\end{table}

\textbf{Zero-shot Capability: }We measure the efficacy of the \BP{} image encoder for describing unseen gestures, i.e., zero-shot generalization capability. This is done by selectively training encoders using only subsets of the gestures:
for JIGSAWS, we test subsets with only gestures 1-5, only gestures 1-10, and no gestures at all (the ``pure CLIP \cite{clip} encoder''), presenting results in Table \ref{tab2_1:zero_shot}.
We disaggregate performance by task, gesture, and model in Table \ref{tab3}.
%Since the JIGSAW tasks do not contain all of the gestures, differing subsets of gestures lead to different numbers of unseen gestures.
The 1-10 gesture model necessarily contains 6 and 8, and thus performance is not reported.
For RARP-45 we test subsets with only gestures 1-3 and gestures 1-6, presenting results in Table \ref{tab2_2:RARP-45}. Per-gesture performance disaggregation is provided as a table in the Appendix. 
For JIGSAWS, we also trained video encoders on one task but evaluated them on the other (``cross-task''), reported at the bottom of Table \ref{tab2_1:zero_shot}. 
This experiment is not possible in RARP-45, as it only has one task. In general we find that \BP{} \OldNewText{pre-training} {prompt-tuning } with only a subset of gestures or even on a different task still provides improvement for most performance measures. This, to us, indicates that \BP{} has relatively high zero-shot capability. %In some cases the reduction in label information increases performance, perhaps by reducing overfit behaviors.

\begin{table}[t]
\caption{\textbf{Above} the double line, we show zero-shot gesture recognition on JIGSAWS with different sets of pre-trained labels (\%), along with an {\color{red} All gestures} supervised case for reference,
and encoders using indices (e.g., ``Gesture 2'') instead of text descriptions during \OldNewText{pre-training}{prompt-tuning} (``ind''). ``No gestures'' has no \OldNewText{pre-training}{prompt-tuning}. \textbf{Below} the double line, we show cross-task zero-shot generalization, using the Knot-Tying \OldNewText{pre-trained}{prompt-tuned} model for Suturing supervised training/evaluation, and vice versa.}\label{tab2_1:zero_shot}
\captionsetup{font=footnotesize}

% \begin{subtable}[t]{\textwidth}
% \subcaption{Zero-shot performance on JIGSAWS Dataset}
% \end{subtable}
\begin{tabular*}{\textwidth}{@{\extracolsep\fill}lcccccc}
\toprule%
& \multicolumn{3}{@{}c@{}}{Knot Tying} & \multicolumn{3}{@{}c@{}}{Suturing} \\
\cmidrule{2-4}\cmidrule{5-7}%
  & Acc. & Edit & F1@\{10, 25, 50\} & Acc. & Edit & F1@\{10, 25, 50\}\\
\midrule
No gestures  & 70.42 & 75.52 & 77.41, 72.52, 59.77 & 75.15 & 78.81 & 83.35, 79.94, 68.02\\ %
5 gestures  & {79.60} & {73.90} & {81.34}, {78.54}, {63.15}  & {81.00} & {82.80} & {87.31}, {85.74}, {76.25}\\ %	
5 gestures (ind)  & {78.93} & {76.10} & {82.63}, {78.82}, {66.30}  & {79.63} & {81.55} & {85.10}, {83.14}, {72.50}\\ %
10 gestures  & {77.07} & 76.56 & 81.79, {77.54}, {66.23} & {81.34} & {82.45} & 87.76, 85.77, 77.86 \\
10 gestures (ind)  & {77.49} & {75.19} & {80.82}, {77.05}, {65.27}  & {81.91} & {82.42} & {87.92}, {85.08}, {77.10}\\ %
{\color{red} All gestures}  & 81.00 & {74.19} & {81.19}, 78.58, 68.31  & 81.72 & 83.89 & {87.34}, {85.57}, {77.61}\\
\midrule
\midrule
Cross-task  & 76.60 & 68.90 & 77.60, 74.16, 60.80 & 78.84 & 81.85 & 85.63, 83.58, 73.77\\ %	
%\midrule
%CLIP \tablefootnote{This is the same as ``No gestures''. ``CLIP'' here means raw CLIP visual encoder that does not have further fine-tuning.} & 70.42 & \textbf{75.52} & 77.41, 72.52, 59.77 & 75.15 & 78.81 & 83.35, 79.94, 68.02\\ %
%Within-task \tablefootnote{This is the same as ``All gestures''. For both ``Cross-task'' and ``Within-task'', the validation sets are the same and just the difference of training sets.}  & \textbf{81.00} & {74.19} & \textbf{81.19}, \textbf{78.58}, \textbf{68.31} & \textbf{81.72} & \textbf{83.89} & \textbf{87.34}, \textbf{85.57}, \textbf{77.61}\\ %	
\botrule
\end{tabular*}
\end{table}

%\setcustomtablecounter{2}{2}
\begin{table}[h]
\captionsetup{font=footnotesize}
\caption{Here we show zero-shot surgical gesture recognition on RARP-45 with different sets of pre-trained labels (\%) on RARP-45 dataset, and the same encoders using indices (e.g., ``Gesture 2'') instead of text descriptions during \OldNewText{pre-training}{prompt-tuning } (``ind'').
%using gesture indices (e.g., ``Gesture 1'', ``Gesture 2'') instead of gesture descriptions during \BP{} training.
}\label{tab2_2:RARP-45}
\begin{tabular*}{\textwidth}{@{\extracolsep\fill}lccccc}
\toprule%
  & Acc. & Edit & F1@10 & F1@25 & F1@50 \\
\midrule
%No gestures (same as ``CLIP'')  & 70.00 & 73.88 & 74.94 & 73.30 & 59.95 \\ %
3 gestures  & 77.78 & 75.09 & 80.96 & 79.28 & 70.60 \\ %		
3 gestures (ind)  & 78.93 & 75.80 & 80.82 & 79.38 & 69.78 \\ %
6 gestures  & 79.19 & 71.85 & 76.60 & 73.68 & 65.17 \\ %
6 gestures (ind)  & 79.73 & 76.51 & 80.70 & 79.06 & 69.01 \\ 
%All gestures (same as ``BP-ViT'')  & 77.36 & 72.16 & 76.29 & 74.50 & 65.32 \\ %
\botrule
\end{tabular*}
\end{table}

\begin{table}[h]
\captionsetup{font=footnotesize}
\caption{We show zero shot F1@10 scores for single gestures on JIGSAWS with different sizes of pre-training sets (\%) that \textbf{exclude} those gestures, along with an {\color{red} All gestures} supervised case for reference, and encoders using indices (e.g., ``Gesture 2'') instead of text descriptions during \OldNewText{pre-training}{prompt-tuning } (``ind''). }\label{tab3}
\begin{tabular*}{\textwidth}{@{\extracolsep\fill}lcccccccc}
\toprule%
& \multicolumn{5}{@{}c@{}}{Knot Tying} & \multicolumn{3}{@{}c@{}}{Suturing } \\\cmidrule{2-6}\cmidrule{7-9}%
  & G11 & G12 & G13 & G14 & G15 & G6 & G8 & G11 \\
\midrule
No gestures  & 89.75 & 65.31 & 59.47 & 60.07 & 84.29 & 58.23 & 46.27 & 75.90 \\ %	
5 gestures  & 91.31 & 75.76 & {70.69} & {73.39} & {89.92} & 79.07 & 69.46 & 92.18 \\ %		
5 gestures (ind)  & 90.65 & 75.98 & 69.00 & 71.28 & 88.48 & 78.68 & 67.66 & 83.04  \\ %	
10 gestures  & 91.79 & {72.89} & 65.87 & 67.21 & {88.68} & - & - & {89.16} \\ %	
10 gestures (ind)  & 92.75 & 71.70 & 70.59 & 70.44 & 86.84 & - & - & 87.97  \\ %	
{\color{red} All gestures}  & 93.61 & 76.55 & 72.17 & 73.49 & 91.01 & 81.20 & {68.11} & 92.86  \\ %	
%No gestures  & 89.75 & 65.31 & 59.47 & 60.07 & 84.29 & 58.23 & 46.27 & 75.90 \\ %	
%\midrule
%5 gestures (prompt)  & 91.31 & 75.76 & \textbf{70.69} & \textbf{73.39} & \textbf{89.92} & \textbf{79.07} & \textbf{69.46} & \textbf{92.18} \\ %	
%\midrule
%10 gestures (prompt)  & 91.79 & \textbf{72.89} & 65.87 & 67.21 & \textbf{88.68} & - & - & \textbf{89.16} \\ %	
\botrule
\end{tabular*}
\end{table}

\textbf{Text description ablation: }
Finally, we show evidence that text descriptions provide negligible benefit for Bridge-Prompt encodings. We measured the value of gesture text descriptions by ablating them to simple ``Gesture Index'' categorical descriptors. For example, gesture 9 could be described as ``using right hand to help tighten suture'' or as ``Gesture 9''. Results from these experiments are included in Table \ref{tab2_1:zero_shot}, \ref{tab2_2:RARP-45}, and \ref{tab3}.

\section{Conclusion and Discussion}\label{sec5}

% In this paper, we explore the effectiveness of a prompt-based video encoder via \BP{} \cite{bridge-prompt} framework. This encoder generates a promising feature for downstream gesture recognition. Furthermore, this encoder demonstrates robust efficacy in zero-shot learning scenarios. The model based on limited label information (or divergent tasks) performs as closely as the performance of the model based on full information (or primary task).

%In this paper, we explore the effectiveness of a prompt-based video encoder via \BP{} \cite{bridge-prompt} framework. This encoder not only generates a promising feature for downstream gesture recognition but also shows a strong zero-shot learning capability. With this prompt-based video encoder, we explore the chance of domain transfer learning. This can be used to avoid time-consuming retraining for specific surgical gestures (task) because the model trained on current knowledge can zero-shot transfer to unseen gestures (other tasks). Robust and generalizable zero-shot learning is a step towards open-set gesture recognition for surgical tasks, thus furthering the potential for surgical automation. 

In this paper we have shown that the \BP{} framework provides both cutting edge gesture-prediction performance for the standard within-task paraadigm as well as strong zero-shot performance on unseen gestures. We believe that this latter case will be essential for any eventual surgical support system. The vocabulary of gestures is too large to learn purely by databases of supervised annotated cases; we should instead plan for situations with weak supervision, and novel gestures in deployment. While the \BP{} framework may not be a component of that eventual system, we believe it makes a significant step towards such a system by demonstrating zero-shot capacity.

%\textbf{Computing resouse limitation}. Batch size and number of frames in one clip are crucial hyperparameters. But increasing it will hugely increase the GPU memory. Hence, limited GPU memory constrains us from further exploration.

%\textbf{Prompt searching}. Altering the prompt is highly likely to influence the ultimate outcome. The process of prompt selection is crucial and empirical. Research on the development of tools to facilitate prompt exploration is increasingly in demand.
\paragraph{Declarations and Acknowledgements}

This research was supported in part by NSF 2321684 and the Wellcome LEAP SAVE program. The authors declare that they have no conflict of interest. All procedures performed in studies involving human participants were in accordance with the ethical standards of the institutional and/or national research committee and with the 1964 Helsinki declaration and its later amendments or comparable ethical standards. This article does not contain any studies with animals performed by any of the authors. This article does not contain patient data.

\bibliography{sn-bibliography}% common bib file

\section{Appendix}
\appendix{
\begin{table}[h]
\caption{Gesture reference table in JIGSAWS dataset.}\label{tab4:gesture_list}
\begin{tabular*}{\textwidth}{@{\extracolsep\fill}cccc}
\toprule%
&Gesture Index & Gesture text description &\\
\midrule
& G1 & Reaching for needle with right hand &\\
& G2 & Positioning needle &\\
& G3 & Pushing needle through tissue &\\
& G4 & Transferring needle from left to right &\\
& G5 & Moving to center with needle in grip &\\
& G6 & Pulling suture with left hand &\\
& G7 & Pulling suture with right hand &\\
& G8 & Orienting needle &\\
& G9 & Using right hand to help tighten suture &\\
& G10 & Loosening more suture &\\
& G11 & Dropping suture at end and moving to end points &\\
& G12 & Reaching for needle with left hand &\\
& G13 & Making C loop around right hand &\\
& G14 & Reaching for needle with right hand &\\
& G15 & Pulling suture with both hands &\\
\botrule
\end{tabular*}
\end{table}

\newpage
% \begin{table}[h]
% \caption{Model performance on RARP-45 dataset and comparisons}\label{tab2}
% \begin{tabular*}{\textwidth}{@{\extracolsep\fill}lccccc}
% \toprule%
%   & Acc. & Edit & F1@10 & F1@25 & F1@50 \\
% \midrule
% All gestures  & 77.36 & 72.16 & 76.29 & 74.50 & 65.32 \\ %	
% 3 gestures  & 77.78 & 75.09 & 80.96 & 79.28 & 70.60 \\ %		
% 3 gestures (index)  & 78.93 & 75.80 & 80.82 & 79.38 & 69.78 \\
% 6 gestures  & 79.19 & 71.85 & 76.60 & 73.68 & 65.17 \\ %
% 6 gestures (index)  & 79.73 & 76.51 & 80.70 & 79.06 & 69.01 \\
% \midrule
% 3D ResNet  & 66.97 & 76.95 & 71.76 & 66.34 & 52.90 \\ 
% I3D  & 65.95 & 74.41 & 71.52 & 65.35 & 51.23 \\ 
% CLIP  & 70.00 & 73.88 & 74.94 & 73.30 & 59.95 \\ 
% \botrule
% \end{tabular*}
% \end{table}

\begin{table}[h]
\caption{Per-gesture performance on RARP-45 dataset (gesture and index)}\label{tab:per_gesture_results}
\begin{tabular*}{\textwidth}{@{\extracolsep\fill}lcccc}
\toprule%
  & G3 & G4 & G7 \\
\midrule
No gestures (CLIP)  & 79.78 & 72.37 & 69.47 \\ % 0.7978	0.7237	0.5869	0.6947
3 gestures  & 84.89 & \textbf{82.56} & \textbf{77.05} \\ %	0.8489	0.8256	0.3282	0.7705
6 gestures  & \textbf{88.85} & - & 76.27 \\ %   0.8885	0.8137	0.3481	0.7627	
All gestures  & 87.44 & 78.69 & 73.38 \\ %	0.8744	0.7869	0.0657	0.7338
\midrule \midrule
3 gestures (ind)  & 85.74 & 84.38 & 73.63 \\ % 0.8574	0.8438	0.4613	0.7363
6 gestures (ind)  & 87.74 & - & 82.17 \\ %0.8774	0.8251	0.4183	0.8217
\botrule
\end{tabular*}
\end{table}

\begin{table}[h]
\caption{Comparison of JIGSAWS and RARP-45 dataset}\label{tab:dataset_comparison}
\begin{tabular*}{\textwidth}{@{\extracolsep\fill}lcccc}
\toprule%
 & & JIGSAWS & RARP-45 & \\
\midrule
& Frame rate (FPS) & 30 & 60 &  \\
& Number of videos & 103 & 36 &  \\
& Average video length (s) & 92.1 & 292.4 &  \\
& Total video length (s) & 9488 & 10526 &  \\
% & Total number of frames & 30 & 60 &  \\
& Total number of gestures & 15 & 7 &  \\
& Task performed & Suturing; Knot tying; Needle passing & Radical prostatectomy &  \\
& Background environment & Lab setting & Operating room &  \\
& Training time (one validation) (h) & 8 & 10 & \\
\botrule
\end{tabular*}
\end{table}

\begin{figure}[h]%
\centering
\includegraphics[width=0.95\textwidth, trim={0.5cm 0.5cm 0.1cm 0.1cm}, clip]{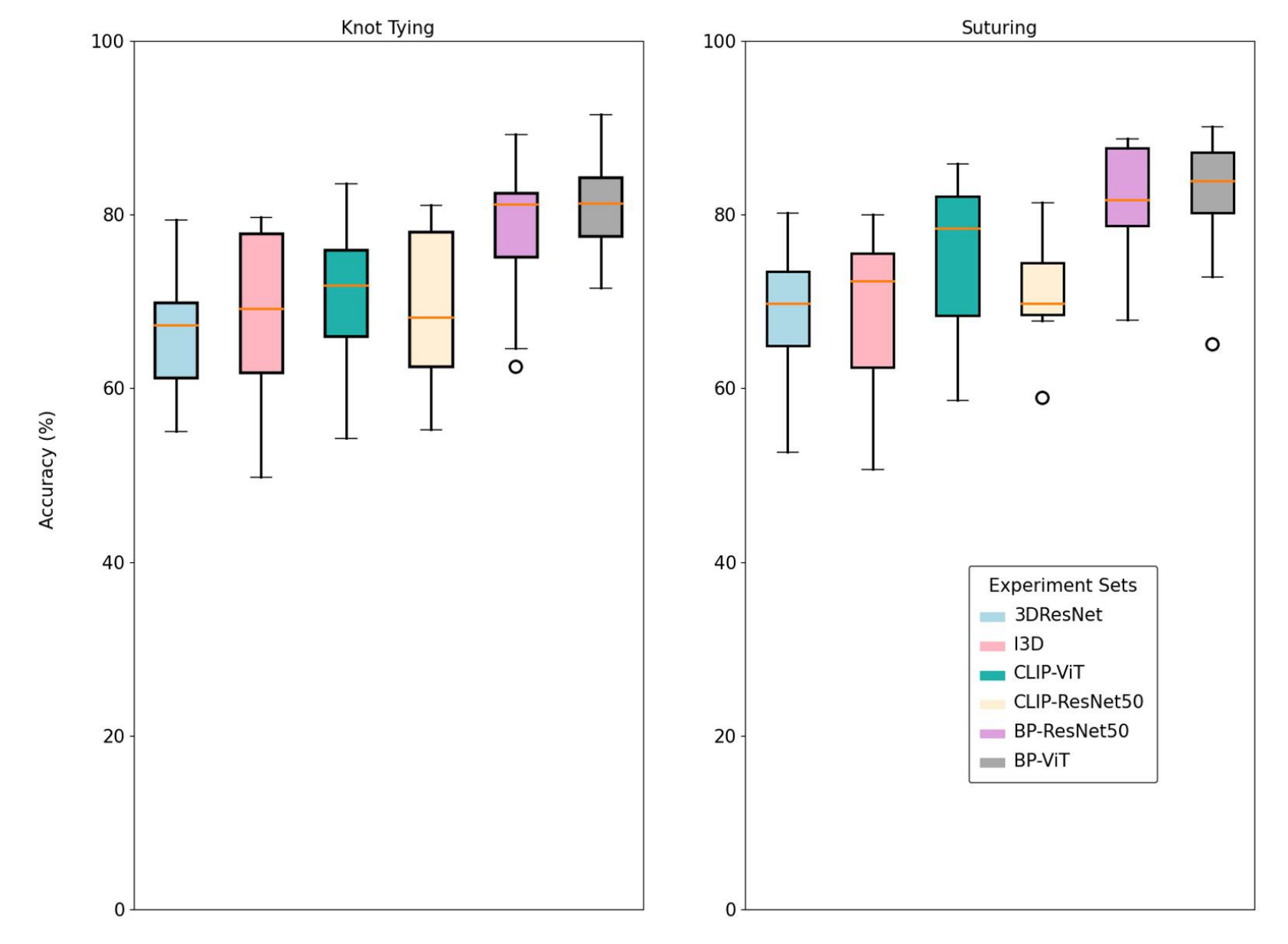}
\caption{JIGSAWS Leave-One-User-Out boxplots for main text table 1. }\label{fig1}
\end{figure} 

\begin{figure}[h]%
\centering
\includegraphics[width=\textwidth, trim={0.1cm 0.1cm 0.1cm 0.1cm}, clip]{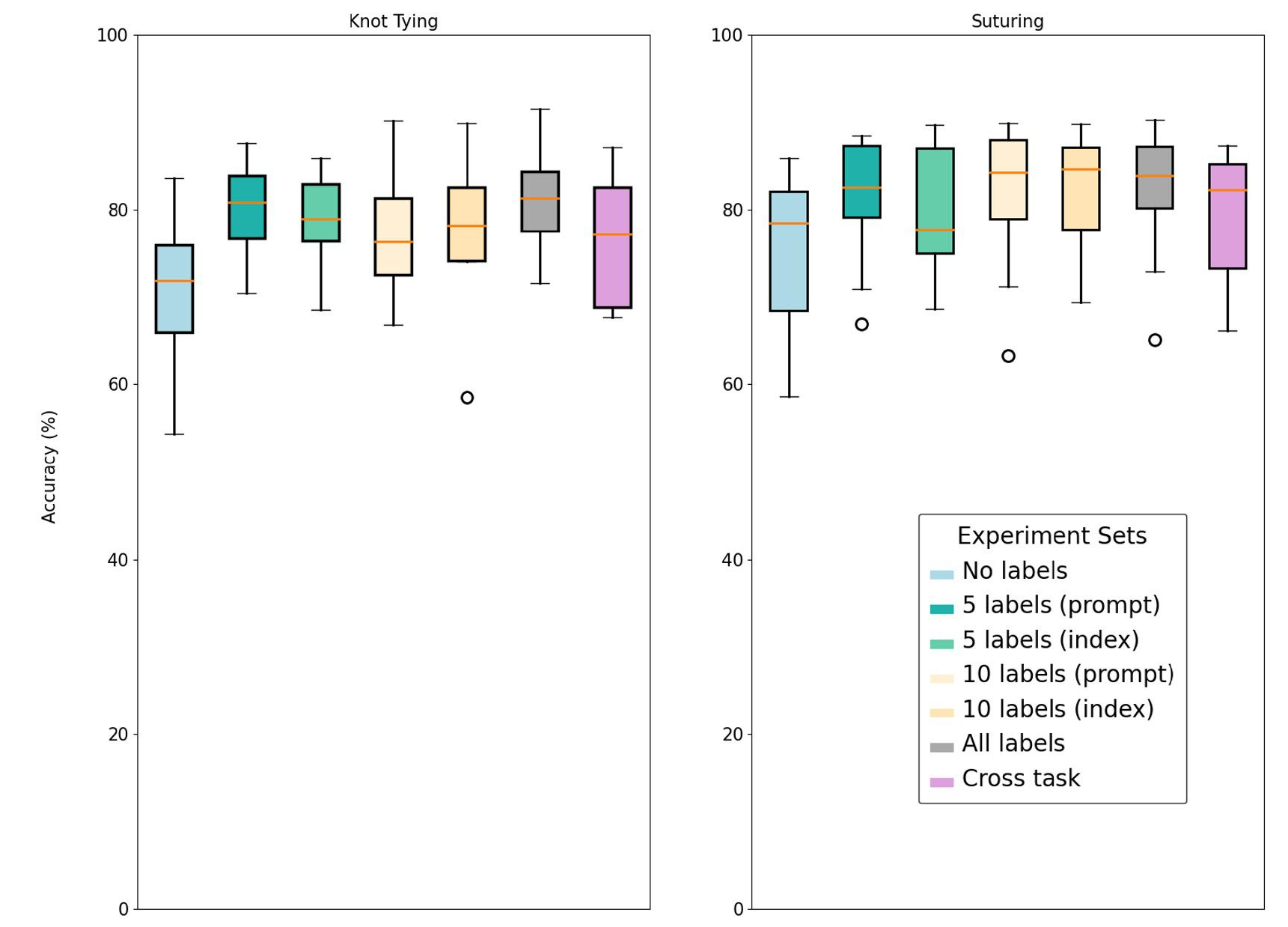}
\caption{JIGSAWS Leave-One-User-Out boxplots for main text table 2. }\label{fig2}
\end{figure} 

\begin{figure}[h]%
\centering
\includegraphics[width=\textwidth, trim={0.1cm 0.1cm 0.1cm 0.1cm}, clip]{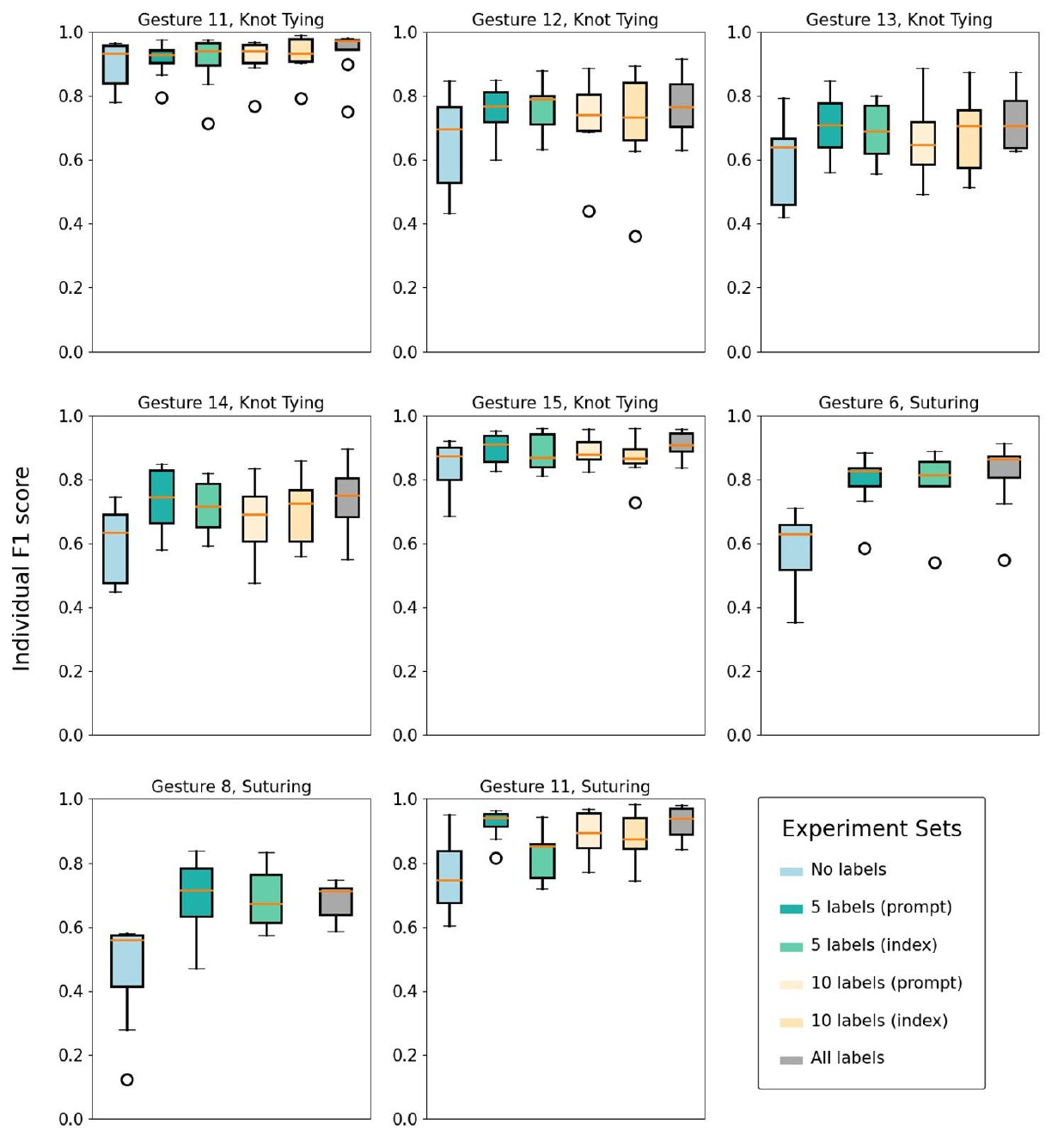}
\caption{JIGSAWS Leave-One-User-Out boxplots for main text table 3. }\label{fig3}
\end{figure} 
}

\end{document}